\documentclass[letterpaper, 10 pt, conference]{ieeeconf}  

\IEEEoverridecommandlockouts                              

\usepackage{amsmath,amssymb,amsfonts}
\usepackage{graphicx}
\usepackage{textcomp}
\usepackage{xcolor}
\usepackage[style=ieee,sorting=none]{biblatex} 
\usepackage{hyperref}
\usepackage{algorithm}
\usepackage{algpseudocode}

\def\BibTeX{{\rm B\kern-.05em{\sc i\kern-.025em b}\kern-.08em
    T\kern-.1667em\lower.7ex\hbox{E}\kern-.125emX}}
\makeatletter
\IEEEtriggercmd{\reset@font\normalfont\fontsize{7.9pt}{8.40pt}\selectfont}
\makeatother
\IEEEtriggeratref{1}

\usepackage{titlesec}

\titlespacing*{\section}
{0pt}{*1.3}{*0.8} 

\titlespacing*{\subsection}
{0pt}{*1.3}{*0.8} 

\title{\LARGE \bf
MambaST: A Plug-and-Play Cross-Spectral Spatial-Temporal Fuser for Efficient Pedestrian Detection
}

\author{Xiangbo Gao$^{1}$, Asiegbu Miracle Kanu-Asiegbu$^{2}$, and Xiaoxiao Du$^{1}$
\thanks{This material is based upon work supported by the National Science Foundation under Grant IIS-2153171-CRII: III: Explainable Multi-Source Data Integration with Uncertainty. A. M. Kanu-Asiegbu is
supported by a Rackham Merit Fellowship.}
\thanks{$^{1}$X. Gao and X. Du are with the Robotics Department, University of Michigan, Ann Arbor, MI 48109 USA
        {\tt\footnotesize xiangbog@umich.edu; xiaodu@umich.edu}}%
\thanks{$^{2}$ A. M. Kanu-Asiegbu is with the Department of Mechanical Engineering, University of Michigan, Ann Arbor, MI 48109 USA
        {\tt\footnotesize akanu@umich.edu}}%
}

\begin{document}

\maketitle
\begin{abstract}
This paper proposes \textit{MambaST}, a plug-and-play cross-spectral spatial-temporal fusion pipeline for efficient pedestrian detection. Several challenges exist for pedestrian detection in autonomous driving applications. First, it is difficult to perform accurate detection using RGB
cameras under dark or low-light conditions. Cross-spectral systems must be developed to integrate complementary information from multiple sensor modalities, such as thermal and visible cameras, to improve the robustness of the detections. Second, pedestrian detection models are latency-sensitive. Efficient and easy-to-scale detection models with fewer parameters are highly desirable for real-time applications such as autonomous driving. Third, pedestrian video data provides spatial-temporal correlations of pedestrian movement. It is beneficial to incorporate temporal as well as spatial information to enhance pedestrian detection. This work leverages recent advances in the state space model (Mamba) and proposes a novel Multi-head Hierarchical Patching and
Aggregation (MHHPA) 
structure to extract both fine-grained
and coarse-grained information from both RGB and thermal imagery. Experimental results show that the proposed MHHPA is an effective and efficient alternative to a Transformer model for cross-spectral pedestrian detection. Our proposed model also achieves superior performance on small-scale pedestrian detection. The  code is available at \href{https://github.com/XiangboGaoBarry/MambaST}{https://github.com/XiangboGaoBarry/MambaST}
\end{abstract}

\section{INTRODUCTION}

Pedestrian detection is an essential task in applications such as autonomous driving. Precise pedestrian detection helps ensure pedestrian safety and helps vehicles to plan paths and avoid collision. Pedestrian detection also has implications in crowd analysis, traffic
monitoring and management, and infrastructure planning \cite{baul2021learning}.   In low-illumination scenarios, such as nighttime, it is
difficult for visible (RGB) cameras alone to detect moving pedestrians. Cross-spectral fusion methods becomes necessary especially under low-light conditions to take advantage of the complementary information provided by both thermal and visible camera data \cite{zhang2023lightweight}. Furthermore, pedestrian video data carries sequential movement information. It is beneficial to incorporate both spatial and temporal information from video frames to enhance pedestrian detection performance \cite{yang2022predicting}.

While significant progress has been made in multi-modal fusion and spatial-temporal modeling, simultaneous cross-spectral
spatial-temporal fusion still lacks exploration. A variety of multi-modal fusion methods have been developed~\cite{liu2016multispectral, li2018multispectral, zhang2019cross, zhang2021guided, chen2021multimodal} for  single-frame cross-spectral spatial fusion. However, these methods are not easily adapted to temporal fusion due to their reliance of 2D image inductive biases—assumptions about spatial relationships and patterns typical of 2D images. For temporal fusion, 3D convolutions~\cite{carreira2017quo, tran2015learning, tran2019video, feichtenhofer2019slowfast}, adaptive 2D convolutions~\cite{huang2021tada, liu2021tam}, and transformers~\cite{yang2022temporally} have been used. However, these methods work for RGB videos and cannot handle multi-modal inputs. 

This paper proposes a novel fusion pipeline that addresses spatial-temporal fusion accounting for cross-spectral (RGB and thermal) sensor inputs. The proposed fusion pipeline, named \textit{MambaST}, is based on a state space model (Mamba) \cite{gu2023mamba}. Mamba is a recent state space model architecture that rivals the classic Transformers \cite{vaswani2017attention} for sequential data processing and has shown initial promise on computer vision tasks \cite{zhu2024vision, zhang2024survey, xu2024survey}.  Our proposed \textit{MambaST} is the first, to our knowledge, that applies Mamba to cross-spectral fusion accounting for both spatial and temporal information. Within \textit{MambaST}, we propose a novel Multi-head Hierarchical Patching and Aggregation (MHHPA) module, which extracts cross-spectral spatial-temporal features across different hierarchical levels. This module is engineered to balance the extraction of fine-grained details with the removal of noise from coarser-grained information. We show that this module can be easily plug-and-play to perform pedestrian detection with YOLO model architecture \cite{redmon2016you} and is an effective alternative to transformer-based modules. We also leverage the recurrent capabilities in the visual state space model~\cite{liu2024vmamba} to enhance the efficiency for \textit{MambaST} in the inference time. We conducted experiments on KAIST, a real-world multispectral pedestrian detection benchmark \cite{hwang2015multispectral}, and we present detailed detection performance evaluation and ablation studies on various parameter choices. Our experimental results show improved pedestrian detection performance and efficiency (e.g., requiring significantly fewer model parameters compared to transformer-based methods).


The contributions of this paper are
summarized as follows.
\begin{itemize}
    \item We propose \textit{MambaST}, a novel cross-spectral spatial-temporal fuser for effective and efficient pedestrian detection. \textit{MambaST} produces superior detection results while requiring less model parameters and GFLOPs.
    \item  We propose a novel plug-and-play MHHPA module for hierarchical spatial-temporal feature extraction.
    \item We show detailed detection performance evaluation and ablation studies on real-world pedestrian dataset.
\end{itemize}

\section{RELATED WORK}

\subsection{Preliminary on Mamba and Vision Mamba} 
Mamba \cite{gu2023mamba} is a recent state space model (SSM) proposed for sequence modeling. It maps input $x(t) \in \mathbb{R}$ to output $y(t) \in \mathbb{R}$ through the following translation formulation
\begin{equation}
\begin{aligned}
\label{eq:lti}
h'(t) &= \mathbf{A}h(t) + \mathbf{B}x(t), \\
y(t) &= \mathbf{W}h'(t),
\end{aligned}
\end{equation}
where $h(t) \in \mathbb{R}^\mathtt{N}$ is the hidden state. $\mathbf{A} \in \mathbb{R}^{\mathtt{N} \times \mathtt{N}}$ is the evolution parameter. $\mathbf{B} \in \mathbb{R}^{\mathtt{N} \times 1}$ and $\mathbf{W} \in \mathbb{R}^{1 \times \mathtt{N}}$ are projection parameters. Originally, Mamba was only used for 1-D sequences. Liu \emph{et al.}\cite{liu2024vmamba} and Zhu \emph{et al.}\cite{zhu2024vision} later adapted the SSM-based model to accommodate 2D image data with two slightly different approaches (named \textit{VMamba} and \textit{Vision Mamba} models). Liu \emph{et al.}\cite{liu2024vmamba} unfolded image patches into sequences along four distinct traversal paths, processing each patch sequence through Mamba, and then merged their outputs. Zhu \emph{et al.}\cite{zhu2024vision} aligned their model architecture with the transformer \cite{vaswani2017attention} architecture and added a positional embedding to each image patch. Both work show the potential of using mamba architecture to extract image features. Dong \emph{et al.}\cite{dong2024fusion}, Li \emph{et al.}\cite{li2024cfmw}, and Peng \emph{et al.}\cite{peng2024fusionmamba} use Mamba for multi-modal fusion but only address single-frame fusion and does not yet generalize to multi-temporal sequences. VideoMamba\cite{li2024videomamba} focuses on temporal fusion but does not address multi-modal spatial fusion. In this work, we build a novel vision mamba-based pipeline for cross-spectral (RGB and thermal) inputs, accounting for spatial and temporal information in video sequences.

\subsection{Cross-Modality Fusion Methods}
Multi-modality sensor data provides complementary information. RGB-Thermal \cite{dasgupta2022spatio, li2023stabilizing, zhang2023lightweight}, RGB-LiDAR \cite{gonzalez2015multiview, el2019rgb,furst2020lrpd}, and RGB-Depth \cite{farahnakian2020rgb,rashed2019motion} are common cross-modality sensor pairings for pedestrian detection in autonomous driving settings. 
Thermal cameras, in particular,  provide finely detailed grayscale images in a variety of lighting and environmental conditions, and are useful sensor sources for fusion, especially in  nighttime and low-light scenarios. A variety of cross-modality (RGB-thermal) fusion methods have been developed based on convolution neural networks \cite{li2018multispectral,zhang2019weakly,zhang2020attention}, probabilistic ensembling methods \cite{chen2022multimodal}, and transformers \cite{qingyun2021cross, lee2022cross, xing2023ms, zhu2023multi}.  Feature fusion has also been used for cross-modality pedestrian detection. For example, Network-in-Network (NIN) \cite{liu2016multispectral} was used to fuse features from different modalities and reduce feature dimensions; INSANet \cite{lee2024insanet} used intra- and inter-spectral attention blocks to learn mutual spectral relationships; and 
Guided Attentive Feature Fusion (GAFF) \cite{zhang2021guided} guided the cross-modal feature fusion with an auxiliary pedestrian mask.

\subsection{Temporal Fusion for Video Understanding}
Fusion methods including 3D convolutions\cite{carreira2017quo, tran2015learning, tran2019video, feichtenhofer2019slowfast}, adaptive 2D convolutions \cite{huang2021tada, liu2021tam}, and Transformers~\cite{yang2022temporally} have been specifically designed for temporal fusion only, but these temporal fusion methods lack the capability of utilizing multi-modal inputs. Other approaches \cite{liu2016multispectral, li2018multispectral, zhang2019cross, zhang2021guided, chen2021multimodal} focus on single-frame cross-spectral spatial fusion and cannot directly adapt to temporal fusion. In this work, we propose to extend a Mamba architecture to account for temporal sequences by recurrently connecting patched feature values across frames.

\begin{figure*}[t]
\includegraphics[width=1\textwidth]{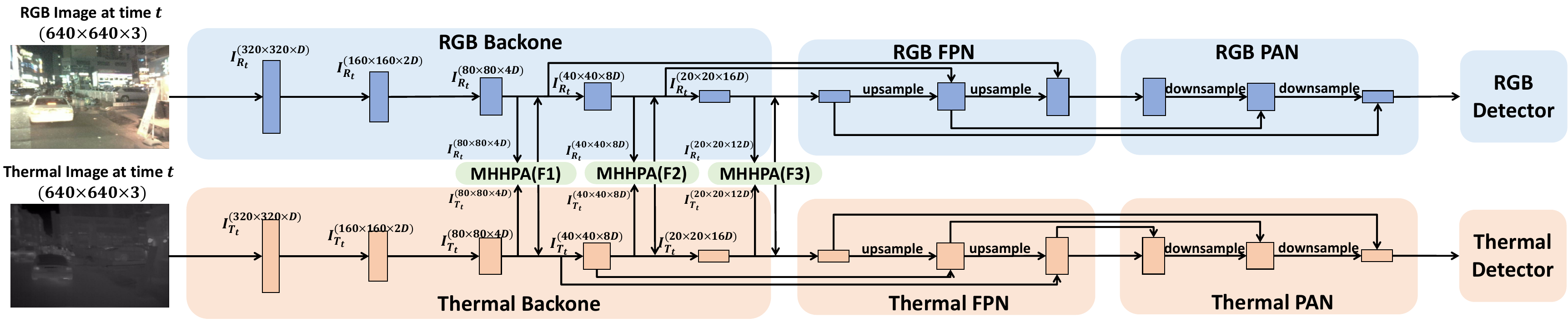}
\vspace{-7mm}
\caption{ \normalsize Visualization of the RGB and thermal object detection network. $D$ denotes the multiplication factor for channel size.}
\label{figure:mambaST_detect}
\vspace{-3mm}
\end{figure*}

\begin{figure*}[t]
\includegraphics[width=\textwidth]{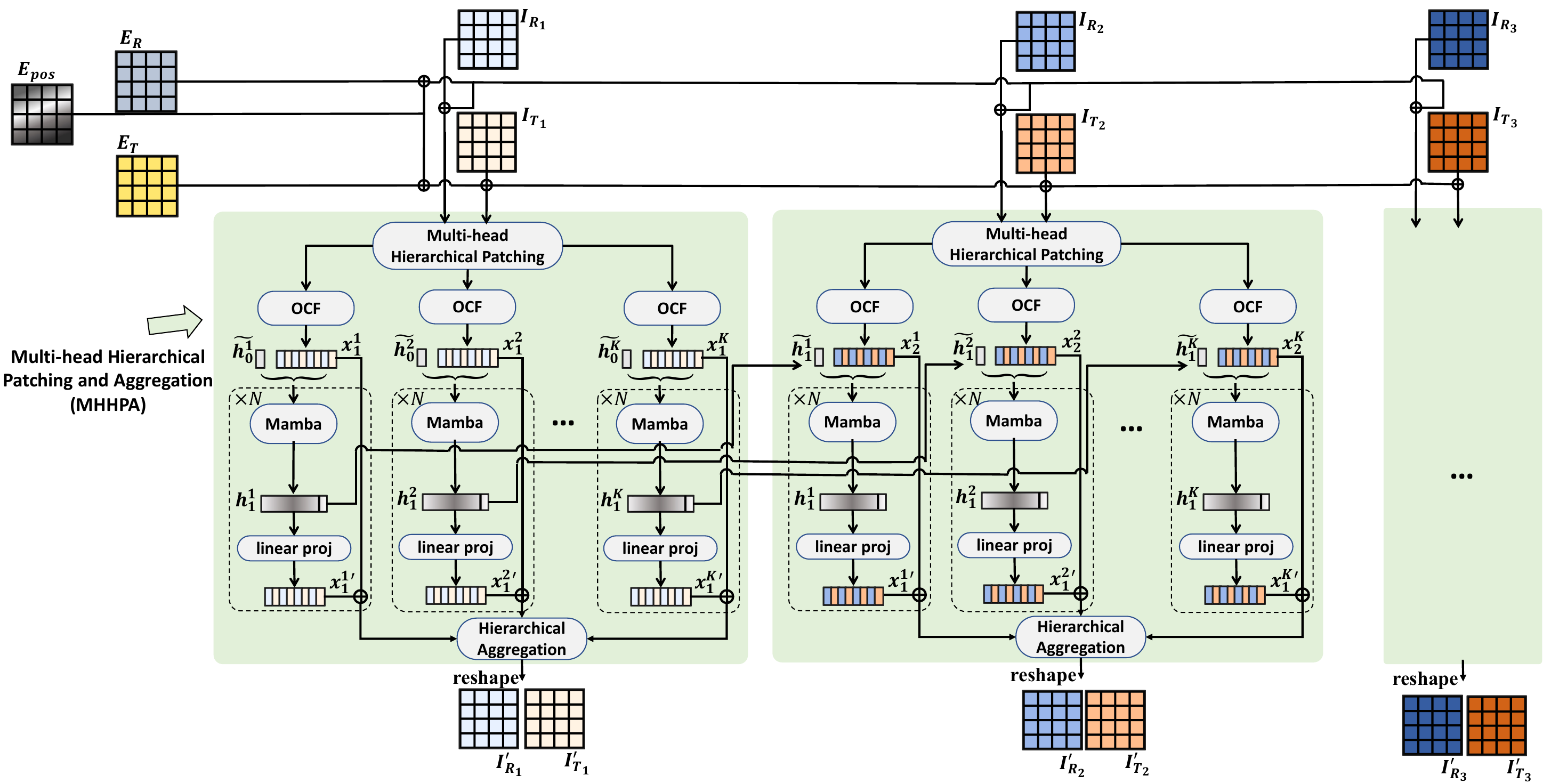}
\vspace{-7mm}
\caption{\normalsize The proposed \textit{MambaST} pipeline. The input RGB and thermal embeddings are passed through a novel Multi-head Hierarchical Patching and Aggregration (MHHPA) module to extract hierarchical features. An order-aware concatenation and
flattening (OCF) procedure is used to concatenate and flatten the patched features.  The MHHPA module was applied recurrently to allow for multi-temporal fusion.}
\label{figure:mambaSTFuser}
\vspace{-5mm}
\end{figure*}

\section{METHODOLOGY}
We propose a novel Mamba-based Spatial Temporal Fuser named \textit{MambaST} for cross-spectral pedestrian detection. The inputs of \textit{MambaST}  are (weakly aligned) multispectral (RGB color and thermal) image pairs containing traffic scenes, including pedestrians. The outputs of \textit{MambaST}  are bounding box detections of pedestrians in each frame.

\subsection{Overview of MambaST Model Architecture}
\label{sec:backbone}
Fig.~\ref{figure:mambaST_detect} illustrates the backbone and object detection pipeline within \textit{MambaST}. We use YOLOv5 backbone, feature pyramid network (FPN) layer, pyramid attention network (PAN) layer \cite{li2018pyramid}, and detector for single-frame RGB and thermal object detection. The RGB and thermal backbone produces \(\mathcal{T}\times 5\times 2\) feature maps. Here, \(\mathcal{T}\) represents the temporal duration, with each modality input yielding five layers of feature maps, and the numeral 2 signifies the two modalities--RGB and thermal. For spatial fusion, let $I_R^{(W_i\times H_i\times C_i)}$ and $I_T^{(W_i\times H_i\times C_i)}$ denote the third, fourth, and fifth layers of RGB and thermal feature maps, respectively, where \((W_i, H_i, C_i) \in \{(80, 80, 4D), (40, 40, 8D), (20, 20, 16D)\}\) are selected and input into the MHHPA (denoted as F1, F2, F3, respectively). Here, $D$ is the multiplication factor for channel size; $W,H,C$ represent the feature map width, height, and channel size, respectively. The output from this module is then added back to the original feature maps, enhancing the fused spatial representation. For notation simplicity, we do not differentiate the notations of feature maps between different fusion layers and use $W, H, C$ to notate the width, height, and channel size for each fusion layer. After the last fusion layer, each feature map is passed into the YOLOv5 FPN layer, PAN layer, and detector for final detection outputs.

\subsection{Input Embedding}
Consider the RGB feature maps \(I_R\in \mathbb{R}^{T\times W\times H\times C}\), and the thermal feature maps \(I_T\in \mathbb{R}^{T\times W\times H\times C}\). We add a positional embedding $E_{pos}$ to each feature map to encode the position information of each feature pixel. We also add thermal embedding $E_{T}$ and RGB embedding $E_R$ for the spectra information of thermal spectrum and RGB spectrum, respectively. All embedding are learnable during training.

\subsection{Multi-head Hierarchical Patching and Aggregation}
We propose a novel Multi-head Hierarchical Patching and Aggregation (MHHPA) structure to extract both fine-grained and coarse-grained information from both RGB and thermal feature maps. Previous work such as VMamba \cite{liu2024vmamba} and vision Mamba \cite{zhu2024vision}, as well as vision transformer \cite{dosovitskiy2020image}, patch and tokenize the feature maps before flattening features, which reduces spatial resolution. This resolution reduction can effectively reduce the time complexity, but can also cause potential information loss and weaken the models' ability to extract fine-grained information. On the other hand, prior work in thermal-RGB fusion such as~\cite{dong2024fusion} flatten the feature map directly for fusion. We claim this is also sub-optimal as, according to Cheng \emph{et al.} \cite{cheng2023towards}, the feature representations of small objects are apt to suffer from  noise. In other words, a highly fine-grained information can be useless or even harmful for feature learning. Thus, in this work, we propose a novel MHHPA structure to simultaneously extract both fine-grained and coarse-grained information through hierarchical structures and then combine them. We show later in the Experiments section that the MHHPA module helps improve detection results.

In the MHHPA module, at the $t^{\text{th}}$ frame, let  \(I_{R_t}\in \mathbb{R}^{H\times W\times C}\) and \(I_{T_t}\in \mathbb{R}^{H \times W\times C}\) denote the RGB and thermal feature maps, respectively. For different patch sizes and given both thermal and RGB feature maps as inputs, the feature maps $I_{T_t}$ and $I_{R_t}$ are firstly patched to $I_{T_t}^k, I_{R_t}^k \in \mathbb{R}^{H/S_k \times W/S_k\times CS_k^2}$, k is the patch size index. Next, they are concatenated and flattened to $z_t^k \in \mathbb{R}^{2HW/{S_k}^2\times CS_k^2}$ following the order-aware concatenation and flattening (OCF) procedure introduced in Sec.~\ref{sec:ocf}. Then, each flattened patches are linearly projected to $x_t^k = z_t^k \mathcal{W}_k$ and passed to the Mamba block to obtain ${x_t^k}' = \text{MambaBlock}({x_t^k}')$. The output from the MambaBlock is reshaped, split, and added back to the patched feature maps to obtain ${I_{R_t}^k}'$ and ${I_{T_t}^k}'$. This procedure will be repeated $N$ times, where $N$ is the number of Mamba layers. Finally, each $I_{R_t}^k$ and $I_{T_t}^k$ are upsampled to their original size and aggregation together by concatenation.

Alg.~\ref{MHHPA} shows the pseudocode of this procedure, where \(\bigcirc\) denotes function aggregation and \(\bigoplus\) represents the concatenation operation over all pixel indices.

\begin{algorithm}[t]
\caption{Multi-head Hierarchical Patching and Aggregation (MHHPA) }
\label{MHHPA}
\begin{algorithmic}[1]
\For{$k = 1$ \textbf{to} $K$}
    \State $I_{R_t}^k \gets \text{Patching}_k(I_{R_t})$
    \State $I_{T_t}^k \gets \text{Patching}_k(I_{T_t})$
    \State $z_t^k \gets \text{OCF}(I_{R_t}^k, I_{T_t}^k)$
    \State $x_t^k \gets z_t^k \mathcal{W}_k$
    \State $h_t^k \gets \bigcirc_i^N \text{MambaBlock}_i(x_t^k)$
    \State ${x_t^k}' \gets \text{Linear}(h_t^k)$
    \State $\tilde{I_{R_t}^k}', \tilde{I_{T_t}^k}' \gets \text{ReshapeSplit}({x_t^k}')$
    \State ${I_{R_t}^k}' = I_{R_t}^k + \tilde{I_{R_t}^k}'$
    \State ${I_{T_t}^k}' = I_{T_t}^k + \tilde{I_{T_t}^k}'$
\EndFor
\State ${I_{T_t}}' = \bigoplus_k \text{Upsample}_k({I_{T_t}^k}')$
\State ${I_{R_t}}' = \bigoplus_k \text{Upsample}_k({I_{R_t}^k}')$
\State \textbf{Output} ${I_{T_t}}', {I_{R_t}}', h_t^{1:K}$
\end{algorithmic}
\end{algorithm}

\begin{algorithm}[t]
\caption{Recurrent Structure for Temporal Fusion}
\label{RNN}
\begin{algorithmic}[1]
\State \textbf{Initialize} $t=1, h_0=0$
\State \textbf{while} $I_{R_t}$ and $I_{T_t}$ exist \textbf{do}
    \State $I_{R_t}', I_{T_t}', h_t^{1:K} = \text{Alg.}~\ref{MHHPA}(I_{R_t}, I_{T_t}, \tilde{h}_{t-1}^{1:K})$
    \State $\tilde{h}_{t}^{1:K} = \text{Last}(h_t^{1:K})$
\State \textbf{end while}
\end{algorithmic}
\end{algorithm}

\subsection{Order-aware Concatenation and Flattening}
\label{sec:ocf}
In the original structured state space sequence model~\cite{gu2023mamba}, the translation function, as defined in Eq.~\ref{eq:lti}, handles consecutive data sequences. To maintain the spatial continuity within multi-spectral feature maps (\(I_R\) for RGB and \(I_T\) for thermal), we introduce an order-aware concatenation and flattening (OCF) procedure. 

Denote a series of patch sizes as $S_{1:K}$ with corresponding projection matrix $\mathcal{W}_{1:K}$, where $\mathcal{W}_k \in \mathbb{R}^{CS_k^2\times C/K}$ is responsible to map channel size to $C/K$ for all head. Here, $K$ is the number of patch sizes. For each frame at time $t$ and patch size index $k$, let \(I_{R_t}^k, I_{T_t}^k \in \mathbb{R}^{H/S_k \times W/S_k\times CS_k^2}\) denote the feature maps for RGB and thermal spectra, respectively. Each pixel within these maps is indexed by its row \(i\) and column \(j\). A pixel at position \((i, j)\) in the RGB and thermal maps is denoted as \(I^k_{{R_t}, (i, j)}\) and \(I^k_{{T_t}, (i, j)}\). The OCF  constructs a feature vector \({x}_t^k\) for the \(t^{\text{th}}\) frame by interleaving pixels from both feature maps, written as
{\small
\begin{equation}
\begin{split}
    & {x}_t^k = \bigoplus_{i} [J^k_{\text{even}, t}(i), J^k_{\text{odd}, t}(i)] \\
    \text{where}\quad & J^k_{\text{even}, t}(i) = \bigoplus_{j\text{ is even}} [I^k_{{R_t}, (i, j)}, I^k_{{T_t}, (i, j)}] \\
    & J^k_{\text{odd}, t}(i) = \bigoplus_{j\text{ is odd}} [I^k_{{R_t}, (i, W/S_k-j)}, I^k_{{T_t}, (i, H/S_k-j)}]
\end{split}
\end{equation}
}

Here, \(\bigoplus\) represents the concatenation operation over all pixel indices \((i, j)\). This approach ensures that the structural integrity and the spatial relationships of the multi-spectral data are maintained in the flattened representation. 

\subsection{Recurrent Structure for Temporal Fusion}
\label{sec:recurrent}

The structured state space sequential model states that the Mamba translation function \eqref{eq:lti} resembles a recurrent neural network structure with an input-variant translation function \cite{gu2023mamba}. To perform temporal fusion, we model recurrent connections between temporal frames on top of the MHHPA module. Suppose our MambaST performed fusion for the first \(t\) frames and produced hidden vector \(h_{t}\in \mathbb{R}^{WH\times C}\). We take the last hidden output, \(\tilde{h}_{t}\in \mathbb{R}^{1\times C}\), concatenate it with the flattened feature map of the \({t+1}^{\text{th}}\) frame \(x_{t+1}\), and input them into \(N\) layers of MambaBlock. This results in the updated outputs \(I'_{T_{t+1}}\) and \(I'_{R_{t+1}}\) by Alg.~\ref{MHHPA}, with \(\tilde{h}_{t+1}\) ready to concatenate to the order-aware flattened feature map of the \((t+2)^{\text{th}}\) frame. The procedure is depicted in Fig.~\ref{figure:mambaSTFuser}, and formulated in Alg.~\ref{RNN}.

\section{EXPERIMENTAL RESULTS}

\subsection{Dataset and Evaluation Metric}
We evaluate our proposed \textit{MambaST} approach on the KAIST Multispectral Pedestrian Detection Benchmark dataset \cite{hwang2015multispectral}. 
For training, the sanitized annotations provided by Li. \emph{et. al.}~\cite{li2018multispectral} which includes 41 video series with 7,601 images pairs are used. While testing is performed on 25 video series with 2,252 images featuring (nearly) aligned thermal-RGB pairs that capture traffic scenes in both day and night/low-light environments. 

We provide evaluation results on two settings from the KAIST benchmark, reasonable and reasonable small. The ``reasonable'' setting includes non-occluded and partially-occluded pedestrians taller than 55 pixels, and the ``reasonable small'' setting includes pedestrians between 50 to 75 pixels in height. Both settings use the log-average miss rate (LAMR) over the range of $[10^{-2}, 10^0]$ false positives per image (FPPI). Lower LAMR corresponds to better performance. We also report recall values, where higher recall is desirable (reduces false negative rate). Additionally, to evaluate the efficiency of the algorithm, we report the number of parameters and giga floating point
operations (GFLOPs) during inference, where lower number of parameters and lower GFLOPs correspond to smaller number of parameters and floating point operations required for processing the image sequences (lower is regarded as more efficient). 

\subsection{Implementation Details}
 The Multi-head Hierarchical Patching and Aggregation (MHHPA) module employs patch sizes $S^{(1)}_{1:4} = \{1,2,4,8\}$ for the first MHHPA block and omits patching for the subsequent blocks. Patch sizes are constrained to powers of two for dimensional consistency. The number of MambaBlock layer $N=8$. For the backbone, we follow the standard YOLOv5L setting and set $D=64$. The number of frames (temporal duration) \(\mathcal{T}=3\), unless otherwise specified (in ablation studies). The KAIST images are of size $640\times512$, and we pad it to $640\times640$ (i.e., $H=W=640$) during training. The original KAIST dataset was captured at 20Hz. To avoid redundancy from consecutive frames, we applied a temporal stride of three, i.e., skipping every two frames. Our proposed network was implemented using Python 3.10.13 and Pytorch 2.1.2, and executed on NVIDIA A100 GPUs.

\begin{table*}[ht]
\centering
\caption{\normalsize Pedestrian Detection Results on the KAIST dataset (full and small).}
\vspace{-3mm}
\label{cmp_small_obj_kaist}
\resizebox{\textwidth}{!}{
\begin{tabular}{l|ccc|c||ccc|c}
\hline
 & \multicolumn{3}{c|}{\textbf{LAMR(\%)$\downarrow$}} &  \textbf{Recall(\%)$\uparrow$} & \multicolumn{3}{c|}{\textbf{LAMR(\%)$\downarrow$}} &  \textbf{Recall(\%)$\uparrow$}\\
\textbf{Fusion Method} & \textbf{All} & \textbf{Day} & \textbf{Night} & \textbf{All} & \textbf{All-Small} & \textbf{Day-Small} & \textbf{Night-Small} & \textbf{All-Small} \\ 
\hline
RGB only & 13.96 & 16.72 & 12.47 & 99.61 & 18.11 & 18.48 & 19.31 & 99.01 \\
Thermal only & 15.54 & 19.64 & 8.28 & 99.52 & 20.67 & 22.19 & 18.59 & 99.20 \\ \hline
Feat. Add. & 12.47 & 15.31 & 7.48 & 99.18 &16.61&19.02&11.80  &98.01\\
CFT~\cite{qingyun2021cross} & 11.34 & 13.37 & 7.26 & 98.42 &16.76 &18.88 &12.43 &96.87 \\
T-CFT & 10.38 & 12.16 & 7.15 & 97.73 &16.11 &17.68 & 12.06 &96.49 \\
D-CFT & 8.68 & 11.45 & 4.53 & 98.69 &15.21 &16.51 &12.97 & 96.59\\
\hline
\textbf{MambaST (Ours)} & \textbf{6.67} & \textbf{8.67} & \textbf{3.12} & \textbf{99.86} &\textbf{11.37} &\textbf{13.56} &\textbf{7.32} & \textbf{99.34}\\
\hline
\end{tabular}}
\label{tab:mr_recall}
\vspace{-3mm}
\end{table*}

\begin{table*}[ht]
\centering
\caption{\normalsize Efficiency Comparison Between Our MambaST and CFT variants on the KAIST Dataset.  }
\vspace{-3mm}
\resizebox{\textwidth}{!}{
\begin{tabular}{l|cccc|cccc|cccc}
\hline
  & \multicolumn{4}{c|}{\textbf{Param.(M)$\downarrow$}} & \multicolumn{4}{c}{\textbf{GFLOPs}$\downarrow$} & \multicolumn{4}{c}{\textbf{Latency (ms)}$\downarrow$}\\
&\textbf{F1} & \textbf{F2} & \textbf{F3} & \textbf{Total} & \textbf{F1} & \textbf{F2} & \textbf{F3} & \textbf{Total} &\textbf{F1} & \textbf{F2} & \textbf{F3} & \textbf{Total}   \\
\hline
CFT~\cite{qingyun2021cross}  & 5.12 & 25.29 & 100.89 & 131.3 & 4.86 & 19.35 & 77.37 & 101.61 &\textbf{8.7} &{8.2} &{8.8} &\textbf{25.6}\\
T-CFT  & 6.42 & 25.41 & 101.17 & 133.01  & 4.87 & 19.37 & 77.39 & 101.63  & 8.9&8.3&8.8&26.0 \\
T-CFT (w/o DS) &16.15 & 30.13 & 103.23 & 149.51 & 2907.72  & 2903.4  & 2901.24  & 8712.36 &570.0&532.0&562.8&1664.8   \\
D-CFT & \textbf{2.86} & 9.62 & 34.94 & 47.42 & 90.6 & 84.93 & 82.1 & 257.63 &20.9&20.7&20.4&62.0 \\
\hline
\textbf{MambaST (Ours)} & \underline{3.07} & \textbf{3.80} & \textbf{15.64} & \textbf{22.52} & \textbf{1.83} & \textbf{1.82} & \textbf{1.79} & \textbf{5.43} &25.8&\textbf{6.8}&\textbf{6.5}&39.1\\
\hline
\end{tabular}}
\label{tab:effciency}
\vspace{-2mm}
\end{table*}

\begin{table*}[ht]
\centering
\caption{\normalsize Ablation study on the KAIST dataset when varying $\mathcal{K}$ (different patch sizes), OCF, and the number of frames $\mathcal{T}$ in training. The median, max, and min values of miss rate were reported across five runs.}
\vspace{-3mm}
\label{cmp_patching}
\resizebox{\textwidth}{!}{
\begin{tabular}{ccc|c|c|ccc|ccc|cccc|cccc}
\hline
\multicolumn{3}{c|}{$K$} & OCF &  $\mathcal{T}$  & \multicolumn{3}{c|}{LAMR (Reasonable) $\downarrow$} & \multicolumn{3}{c|}{LAMR (Reasonable small) $\downarrow$} & \multicolumn{4}{c|}{Para. (M) $\downarrow$} & \multicolumn{4}{c}{GFLOPs $\downarrow$} \\ 

F1 & F2 & F3 &  &  & median & max & min & median & max & min & F1 & F2 & F3 & total & F1 & F2 & F3 & total \\ 
\hline
1 & 1 & 1 &   &3  &  7.22 &	7.86&	6.85    &12.5&	13.08&	12.19 & 8.5 & 16.3 & 55.6 & 80.4 & 2.12 & 1.90 & 1.79 & 5.81 \\ 
2 & 1 & 1 &    & 3 &  7.30&	7.70	&7.10    &	13.68&	14.3&	12.68 &  5.1 & 16.3 & 55.6 & 77.0 & 1.96 & 1.90 & 1.79 & 5.65 \\ 
2 & 2 & 1 &   & 3&  7.19&	7.54&	7.06     &12.39	&12.61	&12.02 &  5.1 & 9.2 & 55.6 & 69.9 & 1.96 & 1.82 & 1.79 & 5.57 \\ 
4 & 1 & 1 &   & 3& \textbf{6.78}&	\textbf{6.95}&	\textbf{6.6}     &\textbf{11.94}&	\textbf{12.84}&\textbf{	11.3} &  4.1 & 16.3 & 55.6 & 75.9 & 1.84 & 1.90 & 1.79 & 5.53 \\ 
4 & 2 & 1 &   & 3&  6.87&	7.58&	6.8     &12.8&	13.15	&11.68&  4.1 & 9.2 & 55.6 & 68.9 & 1.84 & 1.82 & 1.79 & 5.45 \\ 
4 & 2 & 2 &   & 3& 7.48&	7.88&	7.14    &13.72&	14.13&	13.38 &  4.1 & 9.2 & 30.5 & 43.8 & 1.84 & 1.82 & 1.75 & 5.40 \\ 
4 & 4 & 2 &    & 3& 7.81&	8.16	&7.38     &14.21&	14.61&	13.48 &  4.1 & 10.1 & 30.5 & 44.7 & 1.84 & 1.76 & 1.75 & 5.34 \\ 
\hline
4 & 1 & 1 & \checkmark &   1&7.44&	7.82&	6.89 &11.87&13.02& 11.60& 4.1 & 16.3 & 55.6 & 75.9 & 1.84 & 1.90 & 1.79 & 5.53 \\
4 & 1 & 1 & \checkmark &   3&6.73&	7.2&	6.62&11.37&	12.33&	10.92  &  4.1 & 16.3 & 55.6 & 75.9 & 1.84 & 1.90 & 1.79 & 5.53 \\ 
4 & 1 & 1 & \checkmark  &   7&\textbf{6.32}&	\textbf{6.82}&	\textbf{6.20} &\textbf{11.11}&	\textbf{11.68}&	\textbf{10.25} & 4.1 & 16.3 & 55.6 & 75.9 & 1.84 & 1.90 & 1.79 & 5.53 \\  
\hline
\end{tabular}}
\vspace{-2mm}
\end{table*}

\subsection{Comparison with Other Cross-Modal Fusion Methods}
We evaluate our proposed \textit{MambaST} fusion module against the fusion sources (RGB only and Thermal only), as well as a basic feature addition strategy and an advanced Cross-Modality Fusion Transformer (CFT)~\cite{qingyun2021cross}. In the basic feature addition approach, the RGB and thermal features were simply added and the resulting feature maps were broadcasted across modalities. This serves as a baseline comparison. For a more advanced cross-modality fusion approach, we compare to CFT~\cite{qingyun2021cross}, a top-ranking cross modality fuser for pedestrian detection. Note that the original (vanilla) CFT model only works for a single frame. To account for temporal fusion, we implemented three variations of CFT for comprehensive comparison. 1) \textit{CFT} model, where the original CFT was applied frame-to-frame; 2) \textit{T-CFT} model, where the temporal information was integrated by concatenating feature maps from all timesteps, written as 
\begin{equation}
x_{i,k}' = x_{i,k} + \text{CFT}(\text{Concat}_i^N(x_{i,k}));
\end{equation}
and 3) \textit{D-CFT} model, a deformable variant that replaces standard self-attention in the transformers with deformable attention~\cite{xia2022vision} to handle temporal data more efficiently, written as
\begin{equation}
x_{i,k}' = x_{i,k} + \text{DCFT}(\text{Concat}_i^N(x_{i,k})).
\end{equation}


Table~\ref{tab:mr_recall} shows the pedestrian detection results on the KAIST dataset. Overall, the fusion methods outperform the single-modality sources (first two rows), which indicates the necessity of cross-spectral fusion. Thermal-only produced lower miss rate at Night than RGB-only, while RGB-only performed better during the Day. Among the CFT variants, D-CFT (CFT with deformable attention) performed the best compared with temporal concatenation (T-CFT) and original CFT.  Our proposed  \textit{MambaST} outperforms the single-modality baselines as well as all other cross-modal fusion models for both day and night settings on the KAIST dataset, with the lowest log-average miss rate of 6.67\% overall, 8.67\% during the day, 3.12\% at night, and a highest 99.86\% recall. 

\subsection{Evaluation on Small Object Detection}
Following the KAIST benchmark settings, pedestrians beteween 50 to 75 pixels in height are considered small-sized objects. The last four columns in Table \ref{cmp_small_obj_kaist} reports the detection results specifically the small-scaled pedestrians. As shown, the CFT and T-CFT, which use arithmetic addition as a fusion strategy, performed poorly. This is likely due to resolution reduction via average pooling, which removed the fine-grained information for small-scaled objects. Our \textit{MambaST} kept full resolution before the fusion step and produced superior performance across all settings, with a LAMR of 11.37\% overall, 13.56\% during the day, and 7.32\% at night, and achieving the highest recall rate of 99.56\%. This demonstrates the effectiveness of our \textit{MambaST} approach in detecting small-scale pedestrians.

\begin{figure*}[t]
\includegraphics[width=\textwidth]{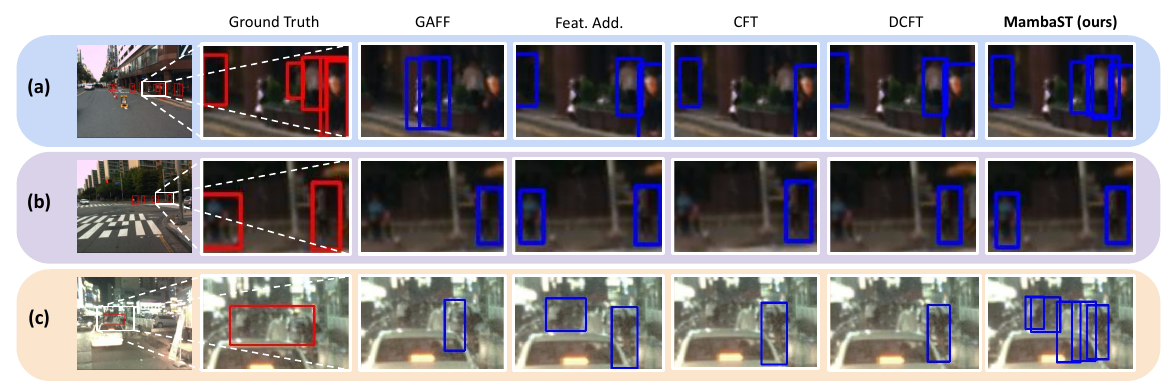}
\vspace{-7mm}
\centering
\caption{\normalsize Visual examples of detection results on the KAIST dataset. All bounding boxes are filtered by confidence $\geq$ 0.5. }
\label{fig:bbox_visual}
\vspace{-3mm}
\end{figure*}

\subsection{Efficiency Evaluation}
Table \ref{tab:effciency} reports the number of parameters, gigafloating point operations (GFLOPs), and latency (ms) during inference. Note that in our T-CFT experiments, the feature maps were first downsampled to $8\times 8$ before fusion. This reduces the number of parameters and produced better results empirically. The T-CFT module, without downsampling (w/o DS), has 149.51M parameters and very high GFLOPs (8712.36). The downscaled T-CFT and the original CFT had significantly lower GFLOPs (101.63 and 33.87, respectively). The deformable variant of CFT (D-CFT) requires fewer parameters but has high GFLOPs. In contrast, our \textit{MambaST} was able to achieve superior detection performance while requiring the lowest number of parameters (22.52M) and GFLOPs (5.43). Our \textit{MambaST} also has a relatively low inference latency (less than 40ms).


\begin{table}[t!]
\centering
\caption{\normalsize Comparison with the state-of-the-art of the KAIST dataset with ``sanitized'' annotation. \textbf{Best}; \underline{Second best} in MambaST performance.}
\vspace{-3mm}
\label{cmp_sota}
\resizebox{0.45\textwidth}{!}{
\begin{tabular}{l|ccc}
\hline
\textbf{Method} & \multicolumn{3}{c}{\textbf{LAMR(\%$\downarrow$)}} \\
& \textbf{All} & \textbf{Day} & \textbf{Night} \\
\hline
MSDS-RCNN~\cite{li2018multispectral} & 7.49 & 8.09 & 5.92 \\
GAFF~\cite{zhang2021guided} & 6.48 & 8.35 & 3.46 \\
MS-DETR~\cite{xing2023ms} & {6.13} & 7.78 & 3.18 \\
CFR~\cite{zhang2020multispectral}&{6.13} & 7.68 &3.19 \\
\hline \hline
MambaST (1 frame) & 7.44& 9.65	& 3.42 \\
MambaST (3 frames) & 6.67& 8.67	& \underline{3.12} \\
MambaST (7 frames) & \underline{6.32} & {8.22}	& \textbf{2.88} \\
\hline
\end{tabular}}
 
\label{tab:sota}
\vspace{-3mm}
\end{table}

\subsection{Comparison with State-of-the-art Methods}
Table~\ref{tab:sota} shows the comparison results against the state-of-the-art fusion methods on the KAIST dataset with ``sanitized'' annotations. MSDS-RCNN~\cite{li2018multispectral} combines a convolution neural network-based multispectral proposal
network and a multispectral classification network to perform fusion. GAFF~\cite{zhang2021guided} proposes inter- and intra-modality attention modules to dynamically weigh and fuse
the multispectral features. MS-DETR~\cite{xing2023ms} fuses  RGB and thermal features in a multi-modal Transformer
decoder and adaptively learns 
 the attention weights. CFR~\cite{zhang2020multispectral} cyclically fuses and refines spectral feature from each modality to achieve cross-spectral complementary/consistency balance. Our \textit{MambaST} achieved competitive detection performance compared with the state-of-the-art methods, and achieves superior detection performance  at Night. As the number of input frames increased, our detection performance also improves, achieving a low 2.88\% LAMR at Night given seven frames as input.

\subsection{Ablation Studies}
We conducted multiple sets of ablation studies to evaluate the effect of parameter choices. 
To reduce the result variance and ensure fairness, we trained using the entire set and selected the checkpoint from the 10th epoch for testing. We also trained the model five times with different seeds for each hyperparameter setup and reported the median overall LAMR.

First, we varied tested different numbers of patch sizes ($\mathcal{K}$) across MHHPA blocks, as outlined in Table \ref{cmp_patching}. The patch sizes range from one size ($\mathcal{K}=1$) to four sizes ($\mathcal{K}=4$) per block, tailored to maintain powers of two for consistent embedding dimensions. The first seven rows in  Table~\ref{cmp_patching} show that the (4,1,1) setting, i.e., using four patch sizes in the first MHHPA block and omitting patching for the subsequent blocks, achieves lowest Log-average Miss Rate (LAMR) in both "reasonable" and "reasonable small" settings of the KAIST dataset without excessive computational overhead. 

Second, we evaluated the impact of the Order-aware Concatenation and Flattening module (OCF). We observed based on row 4 and row 9 in Table~\ref{cmp_patching} that incorporating OCF further enhanced detection performance, reducing the median LAMR from 6.78\% to 6.73\% in the "reasonable" setting and from 11.94\% to 11.37\% in the "reasonable small" setting. 

Third, we performed further experiments varying the number of frames (temporal duration $\mathcal{T}=$1, 3, and 7).  The last three rows in Table~\ref{cmp_patching} show that our model's performance improves with the number of frames used, achieving the lowest LAMR with seven frames as input. This makes sense as one of the advantage of a Mamba-based model is its strength in modeling longer sequences. Future work will include evaluation on longer sequences and other datasets.

\subsection{Visual Results}

Fig.~\ref{fig:bbox_visual} shows example visual results for pedestrian detection on the KAIST dataset. We present our proposed \textit{MambaST} model results, compared to feature addition, CFT, D-CFT, and GAFF models. In row (a), we observed that the CFT model failed to detect some small pedestrians near the center of the scene, while other methods, including the naive feature addition strategy was able to correctly detect more pedestrians. This implies the importance of avoiding resolution reduction in cross-spectral spatial temporal fusion. Similarly, in row (b), our MambaST successfully detected the pedestrians in the scene. Row (c) presents an interesting case, where some  annotations in the KAIST dataset were noisy. It loosely labeled several pedestrians in the same bounding box. As shown, our MambaST model was still able to correctly detect multiple pedestrians in the scene.

\section{CONCLUSION}

We propose \textit{MambaST}, a Mamba-based spatial-temporal fusion pipeline for effective and efficient  cross-spectral pedestrian detection. By utilizing the novel Multi-head Hierarchical Patching and Aggregation (MHHPA) module, \textit{MambaST} efficiently handles the complexities of cross-spectral data without the excessive computational overhead commonly associated with similar models. The MHHPA module can be easily swapped (e.g, with a CFT) and can plug-and-play with a variety of detectors. Our experiments on the KAIST dataset show superior detection performance, particularly in low-light conditions and for small pedestrian detection.
\printbibliography

\end{document}